\pdfoutput=1

\documentclass[11pt]{article}

\usepackage[final]{emnlp2021}

\usepackage{times}
\usepackage{latexsym}

\usepackage[T1]{fontenc}

\usepackage[utf8]{inputenc}

\usepackage{microtype}
\usepackage{tabularx}
\usepackage{graphicx}
\usepackage{float}
\usepackage{makecell}
\usepackage{booktabs}

\title{Generating Related Work}

\author{Darsh J Shah ~~ ~~ ~~
Regina Barzilay\\
Computer Science and Artificial Intelligence Lab, MIT\\
\small{darsh@csail.mit.edu ~~
regina@csail.mit.edu}}
 
\date{}

\begin{document}
\maketitle
\begin{abstract}

Communicating new research ideas involves highlighting similarities and differences with past work. Authors write fluent, often long sections to survey the distinction of a new paper with related work. In this work we model generating related work sections while being cognisant of the motivation behind citing papers. Our content planning model generates a tree of cited papers before a surface realization model lexicalizes this skeleton. Our model outperforms several strong state-of-the-art summarization and multi-document summarization models on generating related work on an ACL Anthology (AA) based dataset which we contribute.
\end{abstract}

\section{Introduction}

\begin{figure}[!t]
\includegraphics[width=1.0\columnwidth]{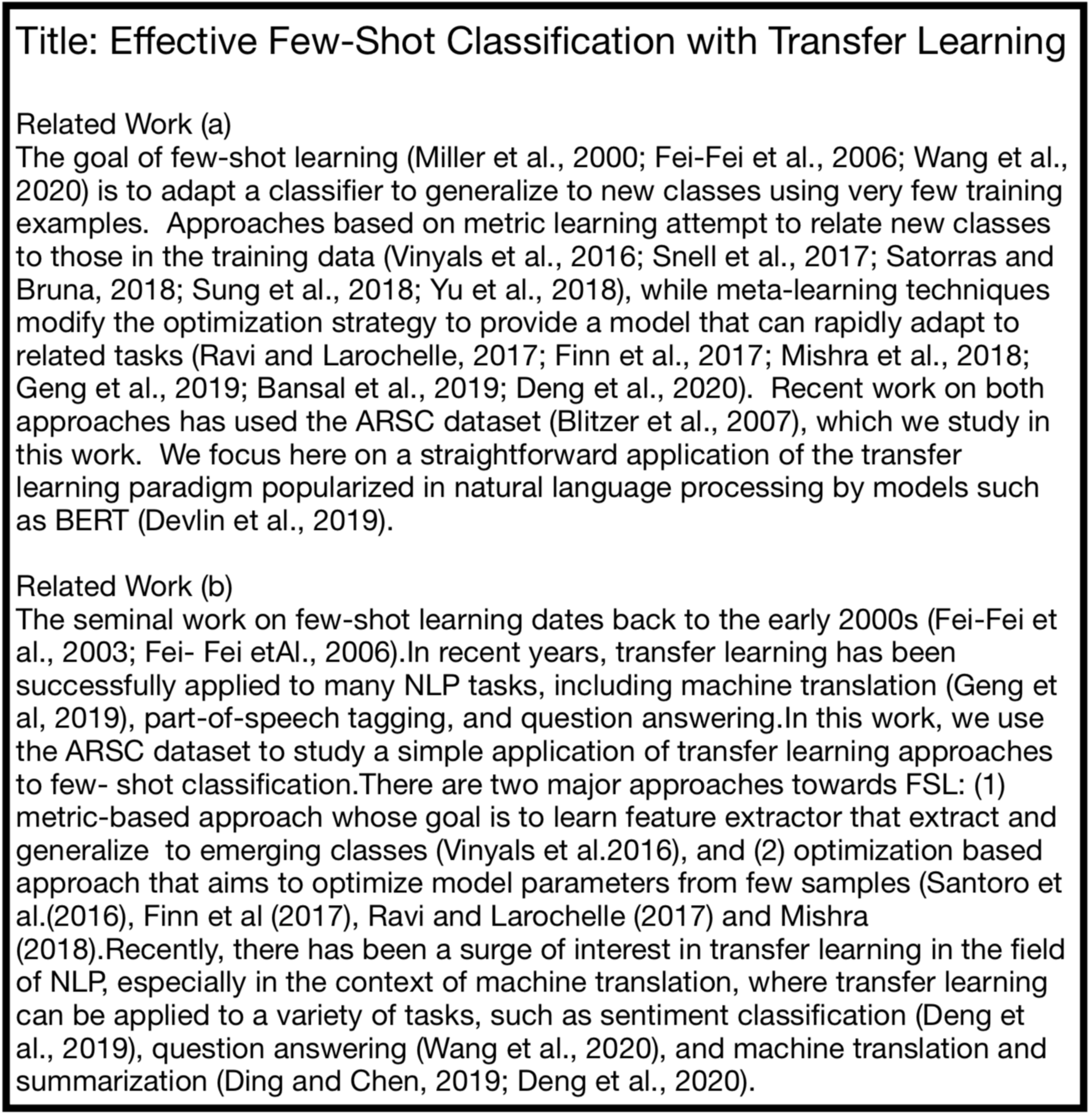}
\caption{Two related work sections presented for the paper \url{https://www.aclweb.org/anthology/2020.coling-main.92/}. Part (a) was produced by the authors and (b) was produced by our model.}
\label{fig:example-1}
\end{figure}

An essential component of scientific writing is positioning new research in the landscape of existing work. Commonly presented in a related work section, this comparison synthesizes information from multiple papers related to the current research.  While identification of such papers can be partially automated\footnote{url{https://www.connectedpapers.com}}, the related work section itself is written by a human writer.  In this paper, we are proposing an algorithm that can assist with this task. Figure \ref{fig:example-1} presents an example where our model generates a related work section for a new paper.

Writing of the related work section is a type of multi-document summarization. However, most of the existing summarization approaches operate over input documents with significant content overlap such as news. These techniques are not applicable to our task since we aim to highlight specific relations of each input article to the current discovery. Prior research in scientific discourse \citep{teufel2002summarizing,teufel-etal-2006-automatic} identified that reasons for citing papers fall into several argumentative classes such as reliance on a previous result or gap in existing solutions.  Therefore, we can view the task of related work generation as predicting such reasons and conveying them in a coherent format.


We implement this approach in a traditional generation pipeline
based on a content planner and surface realizer. Our content planning model organizes the positioning and grouping of past papers predicting the reasons for citing them. The model generates a (depth=2) tree by taking in all available past papers, predicting the  individual reasons for citing them, and producing a sorting of selected papers, grouping them into respective branches with a combined motivation for citing each branch. The produced tree forms the framework to generate readable text. Our surface realization model, which iteratively generates text for every span, fluently lexicalizes the rationale behind citing a set of papers in every branch. Such realization is depicted in Figure \ref{fig:example-2} where a variety of reasons such as similarity (\textit{PSim}), methodological comparisons (\textit{CoCoGM}) or weakness (\textit{Weak}) of past work are captured by our model output. 

Our strategy allows for refined control over the segments, and can appropriately utilize citation categorization annotations available. Specifically, using distant supervision, we can employ the categorizations to learn a pairwise  classification function between a pair of new and old papers -- necessary for generating motivation cognisant outputs. Furthermore, text segmentation annotations on related work sections enable training our content planning and step-wise surface realization models. The approach leaves enough scope for human intervention from an application perspective to modify the skeleton of cited papers or the generated related section.

We apply our method for generating related work on an ACL Anthology dataset that we collect. Every related work section cites multiple scientific studies for varying reasons, making it extremely relevant for our task. We perform extensive automatic and human evaluation to compare our method against state-of-the-art multi-document and query driven summarization techniques. Our approach receives the highest scores on automatic and subjective evaluation metrics such as RougeL, BertScore and Relevance. For instance, on RougeL our approach gets an absolute 5\% over the next best baseline. The method also performs strongly in an update setting, further highlighting its real world applicability. 

\section{Related Work}
{
\begin{figure}[!t]
\center
\includegraphics[width=0.9\columnwidth]{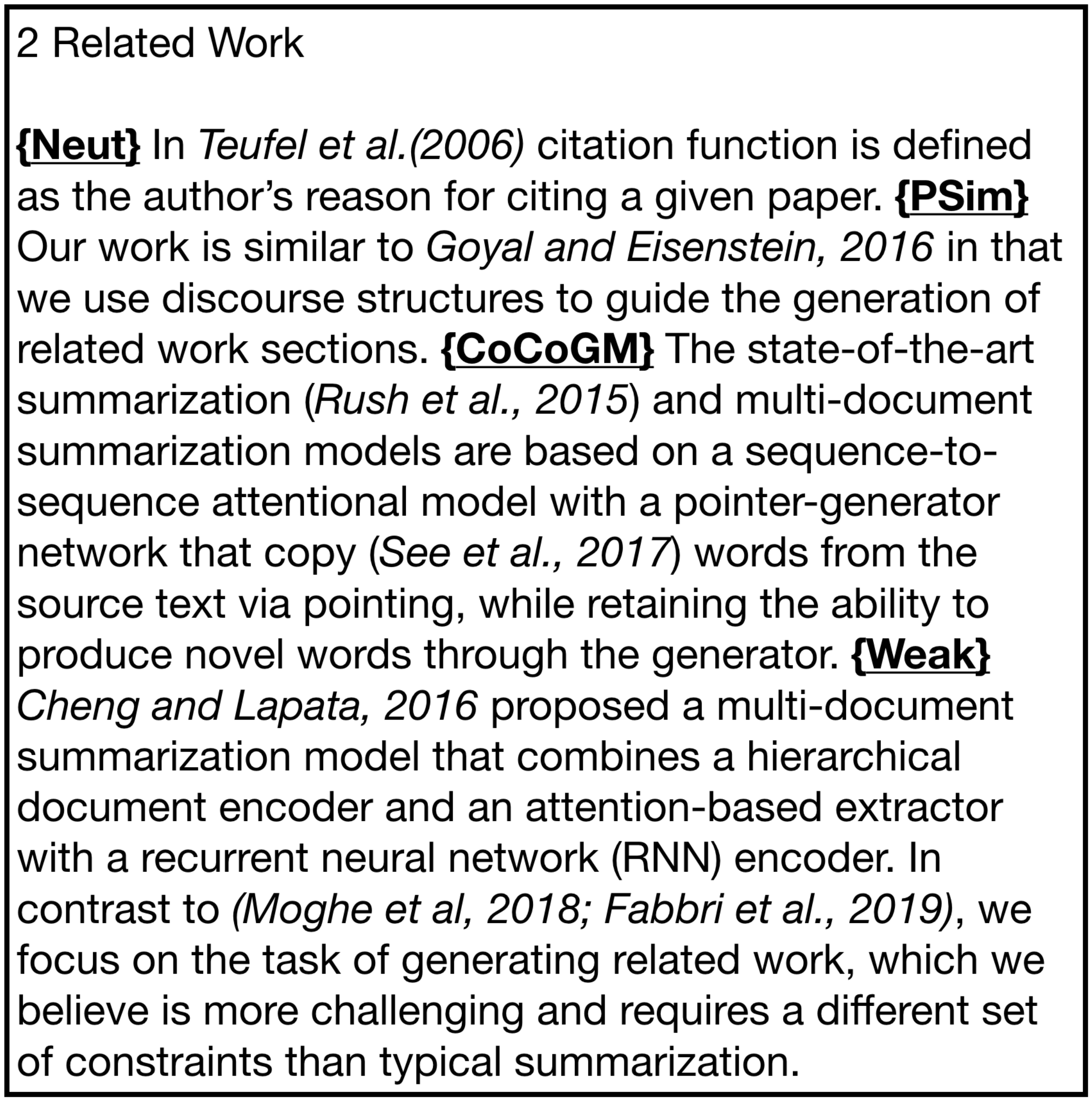}
\caption{Model guided related work produced for the current paper using all ACL Anthology papers which we cite. Our content planner's output branches [\underline{\{reason\}},(cited papers)] guide the paragraph generation by surface realizer.}
\label{fig:example-2}
\end{figure}
}

\paragraph{Multi-document Summarization:} 
Approaches in neural sequence-to-sequence learning~\citep{rush-etal-2015-neural, cheng-lapata-2016-neural,see-etal-2017-get,subramanian2019extractive} for document summarization have shown promise and have been adapted successfully for multi-document summarization~\citep{zhang-etal-2018-adapting, lebanoff2018adapting, baumel2018query, amplayo2019informative, multinews,lu-etal-2020-multi-xscience}. Trained on large amounts of data, these methods have improved upon traditional extractive ~\citep{carbonell1998use, radev-mckeown-1998-generating, haghighi2009exploring} and abstractive approaches~\citep{ mckeown1995generating,ganesan2010opinosis}. The key aspect of typical multi-document summarization solutions is to capture repetitions and similarities in the multiple input documents \citep{barzilay1999information,multinews}. However, in scientific writing, the goal is typically to highlight differences and identify dissimilarities amongst past work. \citet{Shah2021NutribulletsHM} addresses the contrastive nature of scientific studies, however, in related work generation, we are interested in generating an entire section describing past works and their differences in context of the new idea.

\begin{figure*}[t!]
\centering
\includegraphics[width=0.8\textwidth]{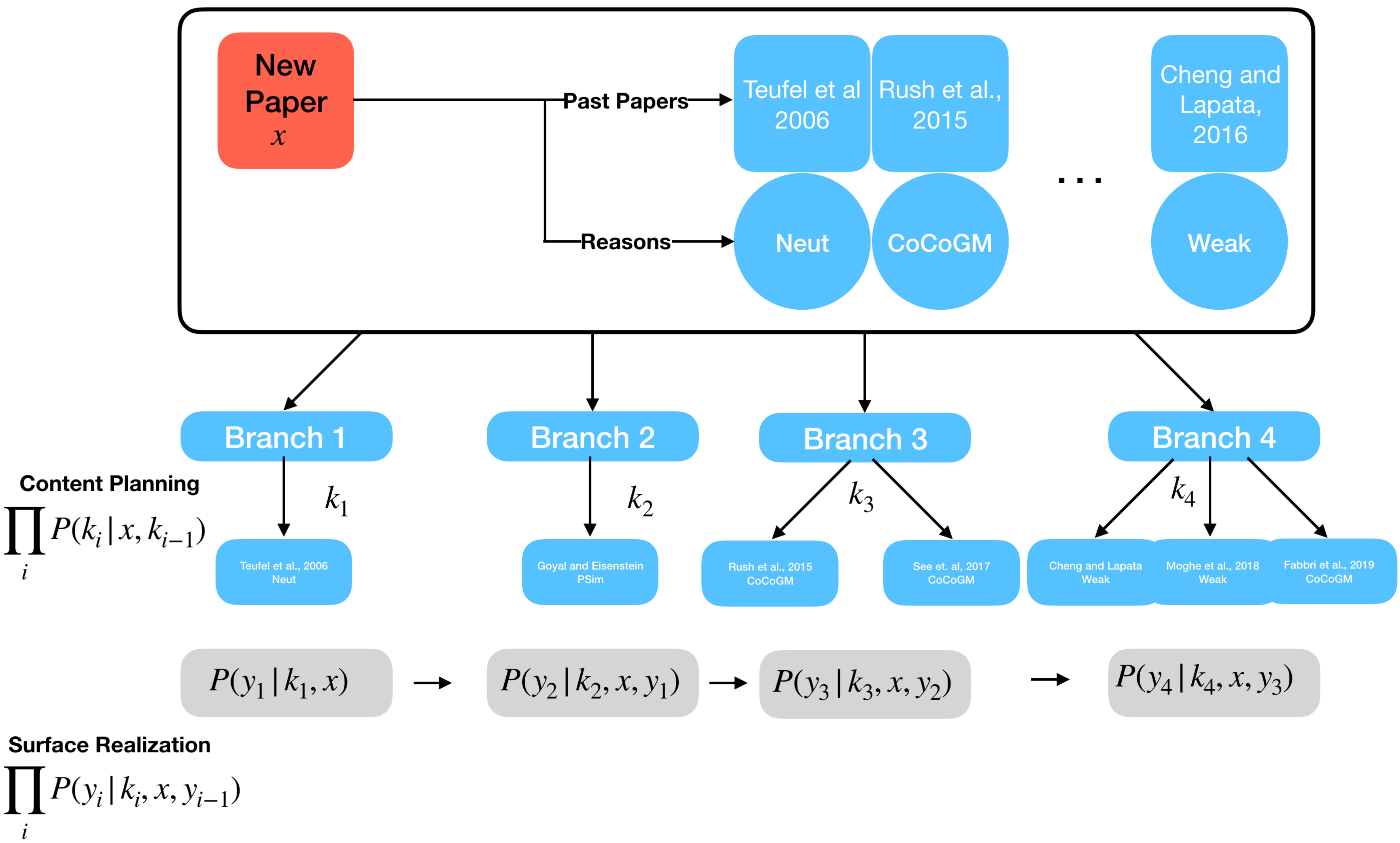}
\caption{Illustrating the flow of our model. $x$ is our paper and the past papers are the ones mentioned in Figure \ref{fig:example-2}. Content planning produces a tree $k$ with four branches with respective citation reasons. Surface realization takes this tree to produce an output $y$.}
\label{fig:model}
\end{figure*}

\paragraph{Query Driven Summarization:} Several tasks such as article generation \citep{46594}, dialogue \citep{moghe-etal-2018-towards,weston-etal-2018-retrieve,dinan2018wizard,fan2020augmenting}, translation \citep{gu2018search} and language modeling \citep{guu2018generating,khandelwal2019generalization} can be categorized as query driven generation. Our work can also be considered as part of this framework, with the new work being a query. However, we are interested in generating a coherent summary which highlights the comparative aspects of past works.
\paragraph{Rhetorical Structure Theory:} Describes the structure of a document in terms of text spans that form discourse units and the relations between them \citep{mann1988rhetorical} and is often used for summarization \citep{goyal-eisenstein-2016-joint,mabona2019neural}. For scientific writing, a flat structure of discourse units rather than a hierarchy has been observed \citep{teufel2002summarizing}. Specifically, for generating related work we base our citation reasoning annotations on \citet{teufel-etal-2006-automatic} to generate informative text. This usage is similar to recent aspect-oriented summarization approaches \citep{kunneman-etal-2018-aspect,frermann-klementiev-2019-inducing}.

\section{Method}

Our goal is to generate a related work section $y$ for a new research idea $x$, using $k$ past papers and rationales to cite them. In this section, we describe our solution, illustrated in Figure \ref{fig:model}. 
\subsection{Overview}
For each paper in the training corpus, we have the abstract $x$, grouping of papers to be cited along with the reasons for citing them $k$ and the related work section to be generated $y$.
Related work sections can be quite long, and in order to model their generation, we break $y$ into segments $\{y_1,y_2,...,y_n\}$ through crowd-source segmentation. Segmentation annotations and subsequent motivation categorization of text give us a grouping for cited papers and reasons for citing them $k=\{k_1,k_2,...,k_n\}$.

The probability of an output summary $y$ is 
\begin{equation}
        P(y|x) = \sum_{k}P(y|k,x)P(k|x)
\end{equation}
\begin{equation}
        P(y|x) = \sum_{k}\prod_{i} P(y_i|k_i,x,y_{i-1}) \prod_{j}P(k_j|x,k_{j-1})
\end{equation}
allowing us to break the problem into two modules of content planning $\prod_j P(k_j|x,k_{j-1})$ and surface realization $\prod_i P(y_i|k_i,x,y_{i-1})$ using Markov assumptions.

The model can be trained and applied in a setting with the gold set of cited papers or a setting where the full set of AA corpus is present and a relevant set must be selected through content planning.

\subsection{Content Planning}
Our content planning model takes a new paper's abstract $x$, a set of available papers to cite and produces a grouped ordering and reason for citing each segment $k=\{k_1,k_2,...,k_n\}$. 
We model our content planning as a tree (depth=2) generation task as depicted in the upper half of Figure \ref{fig:model}. From the \textit{vocabulary} of available past papers, our model generates branches of similar, relevant past papers. The segmentation annotations collected on the related work sections, form our supervision for this generation. For the set of papers to be cited $c=\{c_1,c_2,...c_m\}$ we produce a segmented realization $k=\{k_1,k_2,...,k_n\}$ where $k_i=(\{c_e,...,c_h\},r_i)$, with $\{c_e,...,c_h\}$ the set of papers cited in a branch and $r_i$ the reason for citing them. Table \ref{tab:reasons} mentions a subset of the reasons used in our approach, reported in \citet{teufel-etal-2006-automatic}.

 At every step of our tree generation, the content planning model $\mathcal{F}$ decides whether a particular paper to be cited $c_f$  should be added to the current branch (segment) $k_i$ or to create a new branch $k_{i+1}$. Each paper $c_f$ is represented by the encoding of its title and abstract $e(c_f)$ and the research idea $x$ correspondingly $e(x)$. Tree generation proceeds by considering all yet to be selected past papers and the current sub-tree $b$. The probability that the past paper $c_h$ will be selected for inclusion in $b$ is 
\begin{equation}
\pi_\theta (c_h) = \frac{\exp(\mathcal{F}(b,\hat{x}, \hat{c_h}))}{\sum_{h}\exp(\mathcal{F}(b, \hat{x}, \hat{c_h}))}
\end{equation}Tree generation continues until all papers are used or a maximum number of steps is reached. Each paper $c_f$ has a reason $r_f$ for citing it with respect to the new abstract $x$, predicted through a pair-wise classifier $r_f = f(x,c_f)$. After running the tree generation, each branch $\{c_e,...,c_h\}$ gets a combined reason for citing based on an aggregation function $m$ applied to their individual reasons, $r_i=m(r_e,...,r_h)$.

The generated content planning tree is used by the surface planning model to write a related work section.

\subsection{Surface Realization}
The surface realization model $P(y|x,k)$ performs the critical task of generating a coherent, informative and concise summary $y$ by taking in a new paper's abstract $x$ and the content planning tree $k$ from the previous step as input.

Specifically, we model the long text generation as a step-wise decoding task $P(y|x,k)=\prod P(y_i|k_i,x,y_{i-1})$, with a single segment $y_i$ lexicalized every step. The model is trained using the segmentation annotations on the related work dataset.
The model takes the abstract of new research $x$, abstracts from to be cited papers $\{c_e,...,c_h\}$, a token to represent the reason to cite them $r_i$ and the text span $y_{i-1}$ produced in the previous step. Here, $k_i = (\{c_e,...,c_h\},r_i)$ formulates a multi-document summarization task, controlled by a reason $r_i$. Furthermore, $y_{i-1}$ from the previous step guides the generation of a continuous summary.
The model is implemented using a Transformer \citep{vaswani2017attention} based encoder decoder model. The various inputs $k_i$, $y_{i-1}$ and $x$ are separated using a special token.

\section{Dataset}

\begin{table}[t!]
\center
\scalebox{1.0}{
\begin{tabular}{l|c}
\toprule
Category & Description\\
\midrule
 CoCoXY & \footnotesize{Contrast between two cited methods.} \\
 PSim & \footnotesize{Author's work and cited work are similar.}\\
 Neut & \footnotesize{Neutral description of cited work.}\\
 Weak & \footnotesize{Weakness of cited approach.}\\
\bottomrule
\end{tabular}
}
\caption{Subset of reasons for citing a paper as described in \citep{teufel-etal-2006-automatic}.}
\label{tab:reasons}
\end{table}

In this section, we describe the dataset introduced for our related work generation task. 

\paragraph{AA:} The ACL Anthology (AA) 2020 corpus contains papers on the study of natural language processing and computational linguistics. This corpora covers varied topic areas such as text classification, information extraction, generation, etc. We use this data dump and the corresponding text of the papers, to create our dataset of (paper, author list, title,  abstract, related work section) tuples.
\paragraph{Paper Title and Citation:} We collect the title of the paper and its list of authors. The publication year and author list allow us to identify the acronym used to cite the paper in future works.
\paragraph{Abstract:} We collect the abstract for all papers, which form the description $x$ used to generate related work sections.
\paragraph{Related Work Section:} We parse the related work section for several papers in the AA corpus. These sections contain descriptions of past work, indicated explicitly through acronyms, highlighting the foundations and novelty of new papers. 
\paragraph{Parallel Data:} Considering papers with related work sections and available papers cited, we gather a reasonably sized parallel corpora of 8143 data points split into training, validation and testing sets. All papers published in or before the year 2019 are used for training and 185 and 202 papers from the year 2020 are used for validation and testing respectively. \footnote{Data points considered for evaluation cite at least 15 past papers.}
\paragraph{Segmentation Annotations:} In order to model generating related work, we conduct crowd-sourced text segmentation annotations of the corresponding sections. Annotators are encouraged to identify atomic segments (one or more complete sentences) which describe the same set of related facts.
\paragraph{Citation Reason Annotations:} \citet{teufel-etal-2006-automatic} introduces 26 different reasons for citing past papers (Table \ref{tab:reasons}). We collect the corpora defined in this paper and train a text classification model to identify similar reasons on the current ACL Anthology corpora. Using distant supervision, we use these reasons to collect a pairwise (new paper, cited paper) citation classification corpus.
\section{Experiments}
In this section, we describe the settings used to study the task, evaluation metrics, baselines for comparison and implementation details. 
\paragraph{Settings:} We consider three settings to study the related work generation task. \newline
\textbf{Known Past Works:} We use the gold set of cited papers in order to generate the related work section. \newline
\textbf{Full AA Dataset:} We consider the full set of ACL Anthology papers and expect models to cite some relevant ones to generate a related work section.\newline
\textbf{Related Work Update:} We stimulate a related work modification, where an additional paper is to be cited in an otherwise well written related work section. This scenario is pervasive when authors miss a few references or a new paper is published while writing.

\paragraph{Evaluation Metrics:} We evaluate our systems using the following automatic metrics. \newline
\textit{Rouge} is an n-gram based automatic metric used to compare the model output with the gold reference \citep{lin2004rouge}.\newline
\textit{BertScore} is a contextualized embeddings based automatic metric used to compare the model output with the gold reference \citep{zhang2019bertscore}.\newline
\textit{$k$ Perplexity$^\leftrightarrow$} calculates the perplexity (mean of perplexity and reverse perplexity $\leftrightarrow$) of the reasoning outputs inferred on the model outputs in context of those from human written related work sections in the training data.\newline
\textit{SARI} is a text update evaluation metric \citep{xu-etal-2016-optimizing}, comparing the number of uni-grams added or kept compared to the gold update.
\newline
\newline
In addition to automatic evaluation, we have human annotators score our models on  Relevance and  Fluency for 100 data points per model.\newline
\textit{Relevance:} Indicates if the generated text shares similar information with the reference section.\newline
\textit{Fluency:} Represents if the generated text is grammatically correct and written in well-formed English.\newline
Annotators rate relevance and fluency on a 1-5 likert scale ~\cite{albaum1997likert}. We have 3 annotators score every data point and report the average across the scores.

\begin{table*}[!t]
\small
\centering
\scalebox{0.9}{
\begin{tabular} {l|c|c|c|c|c}
\toprule
\multicolumn{1}{c}{}&\multicolumn{3}{c}{Automatic Evaluation} & \multicolumn{2}{c}{Human Scores}\\
\textsc{Model}           & \textsc{RougeL} & \textsc{Bert-Score} & \textsc{$k$ Perplexity$^\leftrightarrow$} & \textsc{Relevance} & \textsc{Fluency} \\ 
\midrule
Copy-Gen             & 0.17 & 0.60 & 19.8 & 3.47 & \textbf{3.75}\\
Split-Encoder         & 0.19 &0.59 & 18.8 & 3.50 & 3.66\\
MultiDocTransformer  & 0.14 & 0.52 & 19.8 & 3.54 & 3.66\\
TransformerBART          & 0.30 & 0.65 & 12.7 & 3.42 & 3.53\\
\midrule
Ours & \textbf{0.32} & \textbf{0.66} & \textbf{10.5} & \textbf{3.65} & 3.69 \\
\bottomrule
\end{tabular}}
\caption{Evaluation on standard related work generation task.}
\label{tab:automatic-standard}
\end{table*}

\paragraph{Baselines:}
In order to demonstrate the effectiveness of our method, we compare it against several state-of-the-art multi-document summarization methods. \newline
\textit{Copy-Gen:} \citet{see-etal-2017-get} is a summarization technique which can copy from the input or generate words and recently achieved best results on multi-document summarization \citep{lu-etal-2020-multi-xscience}. \newline
\textit{Split-Encoder:} \citet{shah2019automatic} is a double-encoder decoder method for summarization, technique used in query driven summarization and citation text generation tasks \citep{xing-etal-2020-automatic}. \newline
\textit{MultiDocTransformer:} \citet{liu-lapata-2019-hierarchical} is a Transformer \citep{vaswani2017attention} implementation for multi-document summarization.\newline
\textit{TransformerBART:} \citet{lewis-etal-2020-bart} is a pre-trained state-of-the-art summarization \citet{vaswani2017attention} based model.

\paragraph{Implementation Details:} Our tree generation model $\mathcal{F}$ is implemented as a feed forward neural network. We use \citet{lewis-etal-2020-bart} for Surface Realization. We use a BERT \citep{devlin2018bert} based sentence pair classifier for citation reason identification which takes in the abstracts of the new paper and the to be cited paper. In the Full AA Setting models use TfIdf similarity to select the top-$n$ relevant papers to cite.

\section{Results}

In this section, we report the performance of our model and baselines on the standard, AA vocabulary and update settings. We also perform case study and ablation analysis.

\paragraph{Standard Setting:} In this setting, the model is aware of all the papers to be cited while generating the related work. Table \ref{tab:automatic-standard} reports the results of our model and all competing baselines in this setting. Our model scores higher on \textsc{RougeL} and \textsc{BERT-SCORE} than the baselines, generating summaries summaries syntactically and semantically similar to benchmark outputs. \textsc{$k$ PERPLEXITY$^\leftrightarrow$} scores capture realistic variability in reasons for citing papers. Our receives the lowest perplexity score, highlighting its ability to plan and generate a diverse and realistic sequences of citation reasons. 

The \textit{Copy-Gen}, \textit{Split-Encoder}, \textit{MultiDocTransformer} and \textit{TransformerBART} methods focus on the repetitions amongst inputs and fail to bring out detailed similarities and differences necessary in the related work. 
Our method, on the other hand is able to better generate longer sections which are closer to reference summaries and cognisant of the reason for citing papers. 

\begin{table}[!htbp]
\small
\centering
\scalebox{0.9}{
\begin{tabular} {l|c|c} 
\toprule
Model   & RougeL 10 & RougeL 22  \\
\midrule
Copy-Gen & 0.16 & 0.11\\
Split-Encoder & 0.18 & 0.18\\
MultiDocTransformer & 0.24 & 0.24 \\
TransformerBART &  0.24 & 0.24 \\
Ours &  \textbf{0.28} & \textbf{0.29} \\
\bottomrule
\end{tabular}
}
\caption{ Evaluation of full AA database retrieval task.}
\label{tab:full_aa}
\end{table}

\begin{table}[!htbp]
\small
\centering
\scalebox{0.9}{
\begin{tabular} {l|c|c} 
\toprule
Model   & SARI & RougeL  \\
\midrule
Copy-Gen & 0.28 & 0.24\\
TransformerBART &  0.20 & 0.31 \\
MultiDocBART & 0.15 & 0.21 \\
UpdateTransformerBART &  0.15 & 0.22 \\
\midrule
Ours &  \textbf{0.34} & \textbf{0.61} \\
\bottomrule
\end{tabular}
}
\caption{ Evaluation on related work update task.}
\label{tab:updates}
\end{table}

\begin{table}[!htbp]
\small
\centering
\scalebox{0.8}{
\begin{tabular} {l|c|c|c|c} 
\toprule
Model   & Uni-gram & Bi-gram & Tri-gram & Four-gram  \\
\midrule
Copy-Gen & 0.81 & 0.58 & 0.43 & 0.30 \\
Split-Encoder &  0.56 & 0.20 & 0.04 & 0.0 \\
MultiDocTransformer & 0.56 & 0.11 & 0.01 & 0.0\\
TransformerBART &  0.76 & 0.44 & 0.30 & 0.24 \\
\midrule
Ours &  0.78 & 0.39 & 0.21 & 0.15 \\
\bottomrule
\end{tabular}
}
\caption{ Fraction of n-grams copied. }
\label{tab:fraction_copy}
\end{table}

\begin{table}[!htbp]
\small
\centering
\scalebox{0.9}{
\begin{tabular} {l|c} 
\toprule
Model   & Corresponds to Reason \\
\midrule
No-reason input &  0.63 \\
Standard & \textbf{0.65} \\
\bottomrule
\end{tabular}
}
\caption{ Reason ablation. }
\label{tab:reason_ablation}
\end{table}

\begin{table}[!htbp]
\small
\centering
\scalebox{0.9}{
\begin{tabular} {l|c} 
\toprule
Content Planning   & Purity \\
\midrule
K-Means &  0.77 \\
Ours & 0.78 \\
\bottomrule
\end{tabular}
}

\caption{ Content planning clustering purity. }
\label{tab:content_planning_ablation}
\end{table}

\paragraph{Human Evaluation:} Table \ref{tab:automatic-standard} also presents the human evaluation scores of the methods on the related work generation task. Our method is rated the highest by crowd-workers on \textsc{Relevance}, confirming the automatic evaluation metrics. Methods typically produce fluent text. The inter-annotator Kappa agreement \citep{mchugh2012interrater} is 80\% and 83\% for \textsc{Relevance} and \textsc{Fluency} respectively.

\paragraph{Full AA Database Setting:} In this setting, the model is not provided with the gold set of papers to be cited while generating the related work. Our model is trained to select, cluster and order papers for content planning. Table \ref{tab:full_aa} reports our model performance in comparison with baselines from the standard setting. Our tree-segmented content planning model can find a more appropriate ordering of the papers making it easier for content planning to generate reasonable summaries. This allows it to outperform the best baseline by an absolute 5\% on RougeL (in the setting where a maximum of 22 papers are cited). The competing baselines can not deal with the challenges of a growing input and we don't see improvements in their output quality when citing more papers.

\paragraph{Update Setting:} In this setting, an incomplete related work summary is provided to the model. The model must modify the related work (with the data from the already cited works) using the missing cited paper to generate a complete related work. Table \ref{tab:updates} reports the empirical evaluation of our model compared to baselines, including a setting specific \textit{Update-Transformer} method. The flexibility of our tree-segmented content planning approach allows our model to keep most of the existing summary while updating only the requisite segments in a fluent manner to add the missing cited paper. On both SARI and RougeL our model outperforms the baselines by significant margins -- absolute 6\% on \textsc{SARI} and 30\% on \textsc{RougeL}.

\paragraph{Case Study:} Table \ref{tab:casetudy} shows outputs for all the baselines and our model on the paper from Figure \ref{fig:example-1}.\textit{Copy-Gen} cites several papers, but doesn't give an informative summary of their contributions. Alternatively, \textit{Split-Encoder}, while generating a longer section, does not capture the context of the new paper in terms of related work. \textit{MultiDocTransformer} unfortunately doesn't generate a particularly relevant summary for this input. \textit{TransformerBART} generates a paragraph which while being fluent, can't capture the entire pool of related work.  In contrast, our model generates a fluent related work section, covering a lot of the relevant work. Our model also gives insight into the task tackled in the current paper.

\paragraph{Analysis:} We perform analysis to further study the Content Planning and Surface Realization models.

\paragraph{Content Planning:} Our content planning model allows for an organization of past papers cited in the related work generation. The tree-segmentation clusters similar papers and provides the reason for citing each cluster. Table \ref{tab:content_planning_ablation} reports the purity score of the clusters using the tree generation method which is very comparable to a K-Means clustering method that can not even order the segments.

We develop a BERT \citep{devlin2018bert} sentence pair classifier to judge the reason for citing a past paper in context of a new paper. We use \citet{teufel-etal-2006-automatic} citation reasoning annotations as distant supervision to produce training data for sentence pair classification (abstract pairs in this case), achieving an accuracy of 85\% over 26 different classes.

\paragraph{Surface Realization:} Our surface realization model, like baselines, generates a lot of new phrases as reported in Table \ref{tab:fraction_copy}. Our model also produces motivation cognizant outputs (Figure \ref{tab:reason_ablation}). The 2\% improvement compared to no reason input is very significant as the non \textsc{Neut} reasons are less frequent in the test set.

\paragraph{Text Segmentation:} We develop a BERT \citep{devlin2018bert} text segmentation model using MTurk annotations collected on 180 related work sections. This leads to 6000 data points for the binary classification of consecutive sentences into same or different segments. Our model performs this task at a reasonable accuracy of 71\% .

\section{Discussion}
While generating long texts, content control is essential.  This is particularly evident in our study of generating related work which we first model as a tree generation task. This content planning model forms a strong skeleton for coherent and citation reason specific generation. Subsequent lexicalization through the surface realization model produces outputs which outperform those from state-of-the-art methods. This is confirmed by both automatic and human evaluation. 

\paragraph{Human Association in Application:} Separating the solution into planning and generation welcomes human intervention in the writing process. Our reason classification receives an accuracy of 85\% and content planning grouping purity is 78\%. While this performance is indeed promising, the solution also allows for explicit guidance from humans that can improve the final output. Text generation through the surface realization model achieves a state-of-the-art 32\% RougeL F1 score. In a few cases, it may generate phrases which are not factually valid -- prudent human validation is encouraged while using such a system. 

 A human written related work section, on having to accommodate missing papers, would need updating. This forms another scenario for our model to be used to augment an already strong human written summary. Our results in this scenario (Table \ref{tab:updates}) are promising.

Figures \ref{fig:update-example} and \ref{fig:intervention-example} from Appendix \ref{sec:appendix} illustrate examples of an update and human intervention respectively.

\section*{Ethical considerations}
Communicating research ideas is an important task and requires tremendous human effort. While models developed in this paper may be used to develop drafts of related work sections by humans, the primary goal of the work is to model a challenging query driven summarization task in the context of contrastive information from multiple inputs. If our models are applied in the real world, the utmost caution must be applied to ensure no relevant paper is missed or worse, misrepresented. Such mistakes can have a negative impact to a particularly sacrosanct task of surveying fellow researchers' work.

\begin{table*}
\centering
\footnotesize
\scalebox{1.0}{
\begin{tabularx}{\textwidth}{X}
   \toprule
   Copy-Gen \\
   There has been studied in early work such as (Fe-Fei et al., 2003; Fei-Fei et al., 2006) and more recent work (Ba et al., 2016; Santoro et al., 2016; Munkhdalai and Yu, 2017; Ravi and Larochelle, 2017; Ravi and Larochelle, 2016; Mishra et al., 2017; Sung et al., 2018; Allen et al., 2019). More recent work can be divided into two types: similarity-based methods (Vinyals et al., 2016; Snell et al., 2017; Sung et al., 2018) and optimization-based methods (Ravi and Larochelle, 2016; Munkhdalai and Yu, 2017).  \\
    \midrule
\end{tabularx}}
\scalebox{1.0}{
\begin{tabularx}{\textwidth}{X}
   Split-Encoder \\
   Approaches based on metric learning attempt to relate new classes to those in the training data (Vinyals et al., 2016; Snell et al., 2017; Ravi and Larochelle, 2017; Finn et al., 2017; Finn et al., 2017; Vinyals et al., 2016; Snell et al., 2017; Geng et al., 2018; Allen et al., 2018; Allen et al., 2018; Allen et al., 2019). Few-shot text techniques modify the optimization strategy to provide a model to few-shot classes (Vinyals et al., 2014; com/irlab/SmartCiteConBansal et al., 2016; Munkhdalai and Yu, 2017; Finn et al., 2017; Finn et al., 2017; Vinyals et al., 2016; Snell et al., 2017; Geng et al., 2018; Allen et al., 2018; Allen et al., 2018; Allen et al., 2019). The pioneering work in few-shot learning is that of few-shot text is that of informal texts such as discourse-level entailment, or the existence of the shared task (Verhagen et al., 2003; Miller et al., 2003; Wang et al., 2014). Few-shot learning techniques modify the optimization strategy to provide a model that can rapidly adapt to related tasks such as (Fe-Fei (Kim and Larochelle, 2017; Mishra et al., 2018; Yu et al., 2018), while meta-learning techniques modify the optimization (Ba and Larochelle, 2017; Miller et al., 2017). \\
    \midrule
\end{tabularx}}
\scalebox{1.0}{
\begin{tabularx}{\textwidth}{X}
  MultiDocTransformer \\
  -based methods have been proposed for NLP tasks, such as NER (Zhang et al., 2016; Nallapineni and Nallum, 2017; NER, 2017).   Recently, NER have been used for NER has been proposed, including NER and NLP (Zhou and Nenstette, 2017) and NNs (Zeng et alia and Nivreiter, 2015; Nenkova, 2016) have been applied to NER.   In NER tasks, NLP applications such as machine translation (Zampieri et al.g., 2017) (Zadeh et al.(2018) have also been proposed to NMT (Zhao et al, 2017), NER.(2018), NLP community (Zhuber and Nirenburg, 2018; NLP) have shown to NNLP tasks (Zaman and Nadehong, 2018).\\
  \midrule
\end{tabularx}
}
\scalebox{1.0}{
\begin{tabularx}{\textwidth}{X}
  TransformerBART \\
   There are two major approaches towards FSL: (1) metric-based approach whose goal is to learn feature extractor that extract and generalize  to emerging classes (Vinyals et al., 2016; Snell et al.(2017) and (2) optimization based approach that aims to optimize model parameters from few samples (Santoro et al, 2016; Finn etAl., 2017).In this work, we use the ARSC dataset to study a simple application of transfer learning approaches to few-shot classification.We train a single binary classifier to learn all fewshot classes jointly by prefixing class identifiers to the input text.\\
  \midrule
\end{tabularx}
}
\scalebox{1.0}{
\begin{tabularx}{\textwidth}{X}
  Our Method \\
  The seminal work on few-shot learning dates back to the early 2000s (Fei-Fei et al., 2003; Fei- Fei etAl., 2006).In recent years, transfer learning has been successfully applied to many NLP tasks, including machine translation (Geng et al, 2019), part-of-speech tagging, and question answering.In this work, we use the ARSC dataset to study a simple application of transfer learning approaches to few- shot classification.There are two major approaches towards FSL: (1) metric-based approach whose goal is to learn feature extractor that extract and generalize  to emerging classes (Vinyals et al.2016), and (2) optimization based approach that aims to optimize model parameters from few samples (Santoro et al.(2016), Finn et al (2017), Ravi and Larochelle (2017) and Mishra (2018).Recently, there has been a surge of interest in transfer learning in the field of NLP, especially in the context of machine translation, where transfer learning can be applied to a variety of tasks, such as sentiment classification (Deng et al., 2019), question answering (Wang et al., 2020), and machine translation and summarization (Ding and Chen, 2019; Deng et al., 2020).
\\
  \bottomrule
\end{tabularx}}
\caption{Sample outputs for all models considered on the paper from Figure \ref{fig:example-1}. While baselines produce fluent outputs, our model is most appropriate at reporting past work in context of the current problem. }
\label{tab:casetudy}
\end{table*}

\clearpage
\newpage
\newpage
\bibliography{anthology}
\bibliographystyle{acl_natbib}

\appendix

\section*{Appendix}
\label{sec:appendix}
\begin{figure*}[!t]
\center
\includegraphics[width=0.8\textwidth]{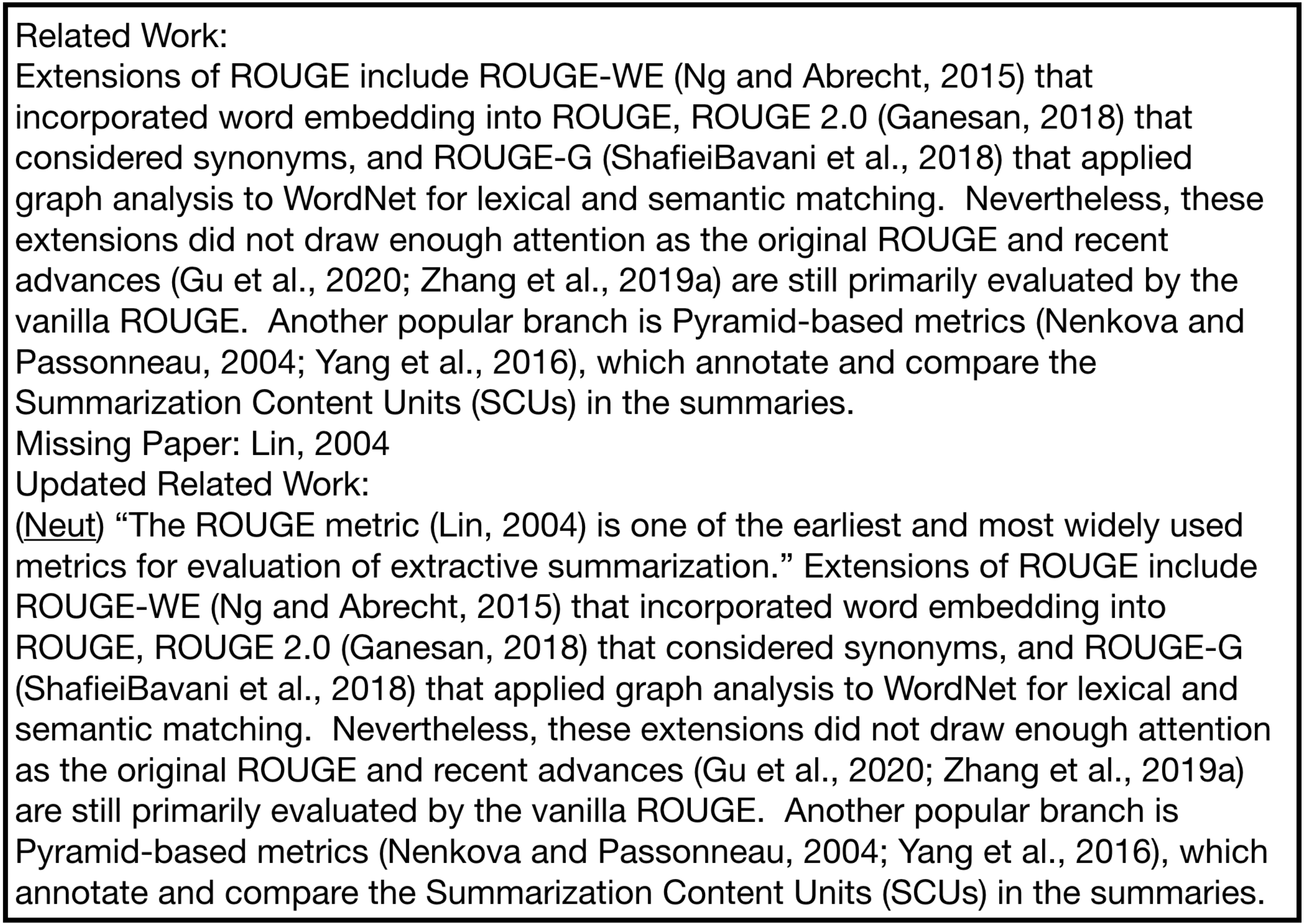}
\caption{Model output for a summary update for an ammortized related work from the paper \url{https://www.aclweb.org/anthology/2020.acl-main.445/}.}
\label{fig:update-example}
\end{figure*}

\paragraph{Update Example:} We demonstrate a scenario where a human written related work missed a critical paper. When provided with the missing paper \citep{lin2004rouge}, our model finds an appropriate branch for the paper and the surface realization completes the writing (Figure \ref{fig:update-example}).

\begin{figure*}[!t]
\center
\includegraphics[width=0.8\textwidth]{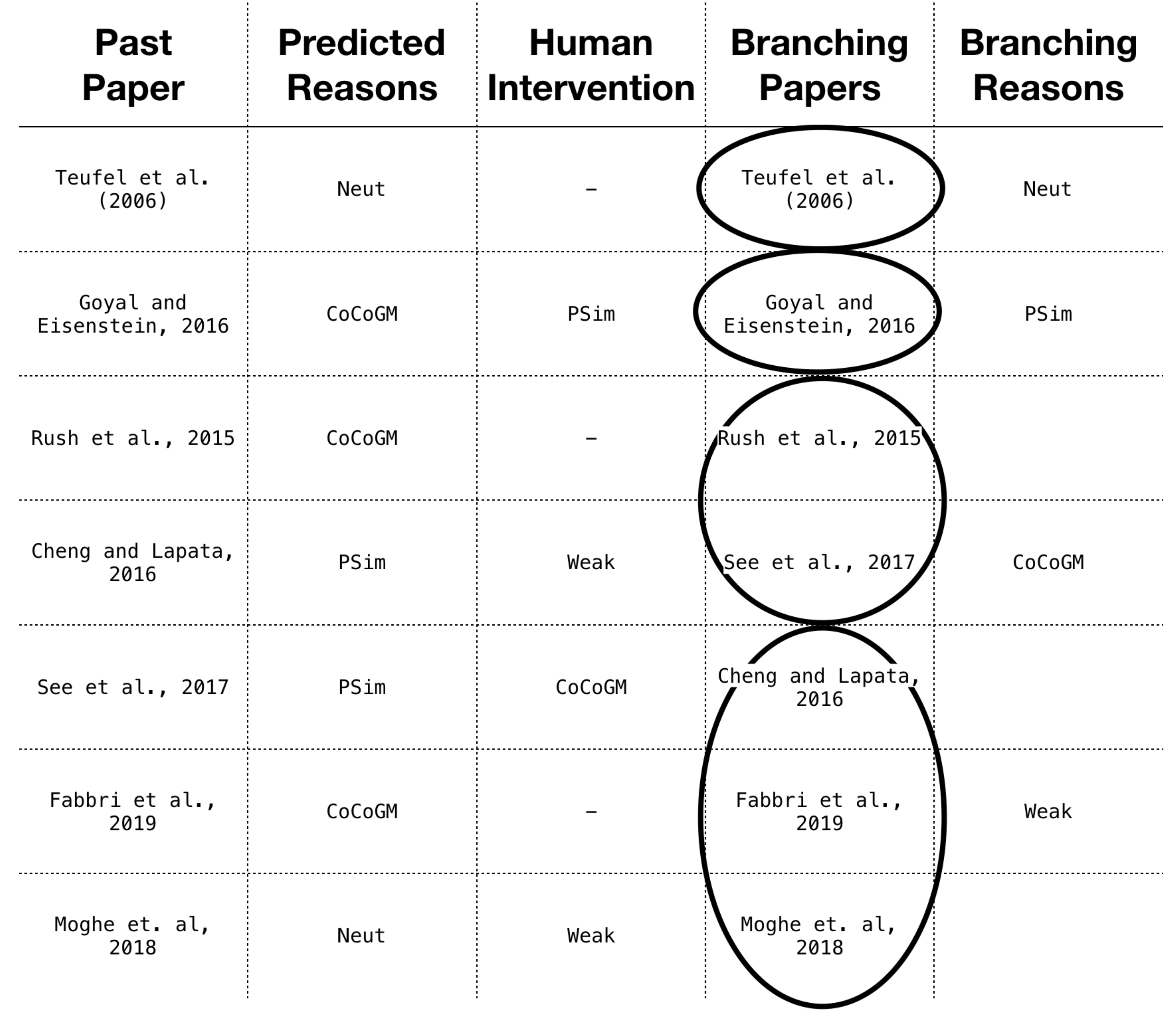}
\caption{Human intervention to produce a more relevant content planning.}
\label{fig:intervention-example}
\end{figure*}

\paragraph{Human Intervention:} The related work generated by our model in Figure \ref{fig:example-2}, we demonstrate how intervention can allow for a more pleasing content planning. Figure \ref{fig:intervention-example} shows the ease with which human intervention can be used to benefit the generation task. The initial reasons for the paper are quite valid, but the modifications allow for a better generation.

\end{document}